\documentclass[journal]{IEEEtran}                           
\usepackage{amsmath,amssymb}                                
\usepackage{amsthm}
\usepackage{psfrag}
\usepackage{epsfig}
\usepackage{graphicx,subfigure}
\usepackage{mathrsfs}
\usepackage{url}
\usepackage{multirow}
\usepackage{booktabs}
\usepackage{bbm,color}
\usepackage[compress]{cite}
\usepackage{mathtools}
\usepackage{array}
\usepackage{supertabular}

\newtheorem{theorem}{Theorem}
\newtheorem{lemma}{Lemma}

\usepackage{algorithm}
\usepackage{algpseudocode}

\begin{document}                                           
\title{\huge A Gegenbauer Neural Network with Regularized Weights Direct Determination for Classification}
\author{Jie He, Tao Chen, Zhijun Zhang, Senior Member, \textit{IEEE}\\
School of Automation Science and Engineering,\\
South China University of Technology\\
Guangzhou, China\\}
\maketitle                                                  
\begin{abstract}                                            
Single-hidden layer feed forward neural networks (SLFNs) are widely used in pattern classification problems, but a huge bottleneck encountered is the slow speed and poor performance of the traditional iterative gradient-based learning algorithms. Although the famous extreme learning machine (ELM) has successfully addressed the problems of slow convergence, it still has computational robustness problems brought by input weights and biases randomly assigned. Thus, in order to overcome the aforementioned problems, in this paper, a novel type neural network based on Gegenbauer orthogonal polynomials, termed as GNN, is constructed and investigated. This model could overcome the computational robustness problems of ELM, while still has comparable structural simplicity and approximation capability. Based on this, we propose a regularized weights direct determination (R-WDD) based on equality-constrained optimization to determine the optimal output weights. The R-WDD tends to minimize the empirical risks and structural risks of the network, thus to lower the risk of over fitting and improve the generalization ability. This leads us to a the final GNN with R-WDD, which is a unified learning mechanism for binary and multi-class classification problems. Finally, as is verified in the various comparison experiments, GNN with R-WDD tends to have comparable (or even better) generalization performances, computational scalability and efficiency, and classification robustness, compared to least square support vector machine (LS-SVM), ELM with Gaussian kernel.
\end{abstract}                                              

\begin{IEEEkeywords}                                        
Extreme learning machine (ELM), least square support vector machine (LS-SVM), regularization neural network, Gegenbauer neural network (GNN), Gegenbauer orthogonal polynomials.
\end{IEEEkeywords}                                          
\section{Introduction}\label{sec.introduction}
\IEEEPARstart{R}{esearches} on artificial feed-forward neural networks (FNNs) have become increasingly active and popular, for it is one of the most powerful tools in artificial intelligence field. What's more, it has been widely studied and applied in a great number of engineering fields, such as the genetic research\cite{liu2014construction}, the prediction of rainfall\cite{nurcahyo2014rainfall}, the analysis of biological emotion\cite{khosrowabadi2014ernn} and the humanoid robots control \cite{Zhang2015Neural}. Because FNN has formidable approximation capability to approximate complex and nonlinear functions directly through training on samples, a computational model containing system characteristics could be built up more conveniently without knowing the exact information of the objective system by training on input samples.

For the development of this algorithm, many researchers have explored the abilities of multi-layer FNNs \cite{MultiLFFNN1,MultiLFFNN2,MultiLFFNN3}.
Leshno has proved that multi-layer FNNs with activation functions are able to approximate any function, including continuous functions and non-sequence functions\cite{MultiLFFNN4}. For the past few years, the single-hidden layer feed-forward neural networks (SLFNs) model has been proposed by Huang \emph{et al.} \cite{ELM001, ELM002, ELM004}, which can approximate functions with zero errors by learning $S$ diverse samples. SLFNs have at most $N$ hidden neurons and the hidden layer biases.

For traditional SLFNs, the parameters need to be tuned iteratively to ensure the optimality of the network and frequent parameters tuning will lead to a complicated relationship between the parameters of different layers. As a solution, the famous Back-propagation (BP) algorithm based on gradient-descend has been proved to be the most effective tool for iterative learning algorithm in SLFNs \cite{Rumelhart1986Learning}. However, the BP-type training algorithm has some inherent weaknesses resulting in inevitable slow operations and poor performances, such as slow convergence \cite{Balakrishnan1992Improving}, local minima \cite{zhang2016survey}.

To address this problem, Huang \emph{et al.} propose and investigate a simple but effective learning algorithm for the SLFNs, called extreme learning machine (ELM). It has been observed that the ELM has unparalleled advantages such as extremely training speed, and excellent generalization performance \cite{ELM001, ELM002, ELM004}. Unlike the traditional BP algorithm, ELM did not pay much attention to the input weights and the neuron biases, and the input weights and hidden layer biases can be arbitrarily assigned \cite{ELM005universal,Huang2012Extreme}.

Although the ELM enjoys its promising merits in application prospects, it still has several shortcomings that ELM could not easily overcome. The first one is that because of due to the mechanisms of randomly assigning weights, the output of ELM fluctuates and the performances could be unstable. The second problem ELM involving is that the hidden layer matrix $H$ could be not full column rank or even ill-conditioned when the input weights and biases are generated randomly, which may incur serious computational robustness problems in obtaining optimal output weights \cite{wang2011study}.

Therefore, in order to overcome the aforementioned weaknesses of BP-type FNNs and ELM, in this paper, we focus on utilizing Gegenbauer orthogonal polynomials as activation functions, and constructing orthogonal polynomials FNN with the simplicity, such as fixed input weights and biases. Such settings have been proved effective in \cite{Zhang2014Weights,JIE2018A}. Further, to avoid lengthy training and determine optimal output weights directly, a method called weights direct determination (WDD) based on pseudo-inverse in previous works \cite{Zhang2014Weights,JIE2018A} is firstly reviewed. However, the WDD only focuses on empirical risk minimization, which might easily lead to a too complicated model and higher risks of over-fitting \cite{Vapnik1995The,Bartlett1998The}. Also, in the previous works, the proposed networks are single output networks, which may be not consistent with the facts that FNNs with multiple output are the most effective way to tackle multi-classification problems.

In this paper, a novel FNN termed as Gegenbauer neural network (GNN) is constructed and investigated based on Gegenbauer polynomials series and polynomials function approximation. Then, based on equality-constrained optimization, a regularized WDD (R-WDD) is proposed with a regularized term to prevent the model from being too complicated. Also, based on the number of network output nodes, a unified learning mechanism of R-WDD for binary and multi classification problems is given. Besides, with regard to computational scalability and efficiency, two different solutions to R-WDD are given, depending on the scale of the problems and the dimensionality of hypothesis feature space. Specifically, the main contributions of this paper lie in the following facts.
\begin{itemize}
\item The main merit ELM algorithm is its extremely fast training speed. In this paper, the proposed GNN has similar calculation complexity to ELM, mainly involving an inversion of a square matrix. Thus, it has the comparable excellent training speed as ELM.
\item Compared with ELM, the proposed algorithm is stable and has lower computational risks, since the input weights are fixed and the tensor products of Gegenbauer orthogonal polynomials are utilized as activation functions.
\item Experiments on several real-world data sets have illustrated that the GNN has comparable or even better generalization performance in binary classification problems and multi-classification problems, compared with least square support vector machine (LS-SVM) with Gaussian kernel and ELM with Gaussian kernel.
\item Experiments have shown that GNN has great computational scalability and efficiency, still performs very well when there only exists very few neurons in the network.
\item Experiments on data sets contaminated with random noises prove the better classification robustness of GNN.
\end{itemize}
The remainder of the article is organized as follows. Section \ref{sec.Theoretical Basis for GNN} presents and analyzes the theoretical basis of the GNN, and the GNN with single output nodes (SOGNN) and the GNN with multiple output nodes (MOGNN) are constructed and discussed respectively. In Section \ref{sec.Proposed Regularized Weights Direct Determination}, the proposed R-WDD is discussed and analyzed based on equality constrained-optimization. Section \ref{sec.experiments and Analysis} presents several comparison experiments on LS-SVM, ELM and GNN with R-WDD. And the final the remarks are given in Section \ref{sec.conclusion}.

\section{Theoretical Basis for GNN}
\label{sec.Theoretical Basis for GNN}
In this section, the fundamentals of Gegenbauer Polynomials are first presented. Then, the theoretical basis of Gegenbauer polynomials series (GPS) for multivariate function approximation is briefly analyzed. Furthermore, based on the analysis, the GNN model is constructed and investigated.
\subsection{Fundamentals of Gegenbauer Polynomials}\label{subsec.Fundamentals of Gegenbauer Polynomials}
\textbf{Definition 1.} The general expression of Gegengenbauer polynomials can be defined as below \cite{Zhang2009Gegenbauer,Yin2017A}:
\begin{equation}
\label{eq.gegenbauergeneral}
g_{n}^{\lambda}(x)=\sum^{n}_{k=0}\frac{1}{k!(n-k)!}\frac{(2\lambda)_{n}(2\lambda+n)_{k}}{(\lambda+1/2)_{k}}(\frac{x-1}{2})^{k},\\
n\in N
\end{equation}
\noindent where $\lambda$ is a parameter of the polynomials that satisfies $\lambda > 0$, $n\in N$ is the degree of the Gegenbauer polynomials and $(\lambda)_{k}=\lambda(\lambda+1)\cdots(\lambda+k-1)=\Gamma(\lambda+k)/\Gamma(\lambda)$ and $\Gamma(\cdot)$ is a Gamma function. Equivalently, Gegenbauer polynomials can also be defined by the recurrence relations below \cite{Yin2017A}:
\begin{equation}\label{eq.gegenbauerrecurrence}
\left\{
\begin{array}{ll}
g_{n}^{\lambda}(x)=1, & n=0  \\
g_{n}^{\lambda}(x)=2\lambda x, & n=1\\
g_{n}^{\lambda}(x)=\alpha_{n} xg_{n-1}^{\lambda}(x)-\beta_{n} g_{n-2}^{\lambda}(x), & n \geqslant 2
\end{array}
\right.
\end{equation}
\noindent where $\alpha_{n}=2(n+\lambda-1)/n$ and $\beta_{n}=(n+2\lambda-2)/n$.

According to orthogonal polynomials theories \cite{Wu2013Interval,Xu2014Uncertainty}, Gegenbauer polynomials are orthogonal with respect to weighted function $\rho (x) \geqslant 0$ within the interval $x \in [0,1]$, and have the orthogonal properties as below:
\begin{equation}
\label{eq.orthogonalfunction}
\begin{aligned}
[g_{i}^{\lambda}(x), g_{j}^{\lambda}(x)]&=\int_{0}^{1} \rho(x)g_{i}^{\lambda}(x)g_{j}^{\lambda}(x)dx\\
&=\left\{
\begin{array}{ll}
0, & i\neq j  \\
\int_{0}^{1} \rho(x)(g_{i}^{\lambda}(x))^{2}dx>0, & i=j
\end{array}
\right.
\end{aligned}
\end{equation}
\noindent where $\rho (x)$ satisfies that $\int_{0}^{1} \rho(x)dx>0$ and there exists $\int_{0}^{1} x^{n}\rho(x)dx > 0, n \in N$.

\subsection{GPS for Multivariate Function Approximation}
\label{subsec.GPS for Multivariate Function Approximation}

First, a function approximation theorem, i.e., Weierstrass approximation theorem, is presented for the further discuss.

\begin{theorem}
\label{approximate}
(Weierstrass Approximation Theorem)\\
Let $\phi \in C\left( U,\mathbb{R} \right)$ with $U\text{=}\left[ a,b \right]$.Then there exists a sequence of polynomials ${{p}_{n}}\left( x \right)$ converging uniformly to $\phi\left( x \right)$ on $\left[ a,b \right]$:

\begin{equation}
\label{weierstrass1}
  \left| \phi-{{p}_{n}}\left( x \right) \right|<\xi
\end{equation}
with $\forall \varepsilon >0$. And it can also be generalized to a multivariate case. Set ${{S}_{m}}=\left\{ \left( {{x}_{1}},\ldots {{x}_{m}} \right)\in {{\mathbb{R}}^{k}}|a<{{x}_{t}}<b,t=1,2,\ldots m \right\}$. Let $\phi\in C\left( {{S}_{m}},\mathbb{R} \right)$. Then there is a sequence of polynomials ${{p}_{{{n}_{1}},\ldots {{n}_{m}}}}\left( {{x}_{1}},\ldots {{x}_{m}} \right)$ converging uniformly to $\phi\left( x \right)$ on ${{S}_{m}}$:

\begin{equation}\label{weierstrass2}
  \left| \phi-{{p}_{{{n}_{1}},\ldots {{n}_{m}}}}\left( {{x}_{1}},\ldots {{x}_{m}} \right) \right|<\xi
\end{equation}
with $\forall \xi >0$.
\end{theorem}
\begin{IEEEproof}
 The proof of famous Weierstrass approximation theorem has been completed in \cite{Giardina2005Proof}. Thus, the proof of Theorem \ref{approximate} is completed.
\end{IEEEproof}
Based on the Gegenbauer polynomials and theories of function approximation, i.e., Theorem \ref{approximate}, and polynomial interpolation \cite{Jie2015Dynamic,JIE2018A,Zhang2008A}, a univariate continuous function $\phi(x)$ defined on $x\in [0,1]$ can be approximated by the $N$-order (or to say the total number of polynomials used) GPS:

\begin{equation}
\label{eq.gegenbauerapproximation}
\begin{aligned}
\phi(x)&= p_{N}(x)+\xi_{\phi}(N)\\
&=\sum_{i=0}^{N}\mu_{i}g_{i}^{\lambda}(x)+\xi_{\phi}(N),
\end{aligned}
\end{equation}

\noindent where $\mu_{i}$ is the coefficient corresponding to $g_{i}^{\lambda}(x)$, $p_{N}(x)$ denotes $N$-order GPS and $\xi_{\phi}(N)$ is the error function related to $N$. According to approximation theories \cite{Zhang2014Weights} and Theorem \ref{approximate}, the error function could converge to zero when $N$ is large enough, which could be expressed as:

\begin{equation}
\label{eq.error-convergence}
\lim_{N\rightarrow+\infty} \xi_{\phi}(N)=0.
\end{equation}

For the function approximation of GPS in $m$-dimension, we have the following lemma \cite{Ibikli2008Direct,D2006Extension}.

\begin{lemma}
\label{lemma.Bernstein}
Let $D=[0,1]^{\rm m\times1}$ and $\phi(x)$ is a continuous real-valued function with $m$ variables on $D$ (i.e., $x=[x_{1}, x_{2},\dots,x_{m}]^{\rm T} \in [0,1]^{\rm m\times1}$). Then, $B_{n}(\phi;x)$, the generalized multivariate Bernstein polynomials of $\phi(x)$, converges uniformly to $f(x)$ and could be structured as follows:
\begin{equation}
\label{eq.Bernstein}
B_{n}(\phi;x)=\sum^{n}_{k_{1}=0}\cdots\sum^{n}_{k_{m}=0}\phi(\frac{k_{1}}{n},\cdots,\frac{k_{m}}{n})\prod^{m}_{t=1}b_{n,k_{t}}(x_{t}),
\end{equation}
\noindent where $b_{n,k_{t}}(x_{t})=C_{n}^{k_{t}}x_{t}^{k_{t}}(1-x_{t})^{n-k_{t}}$ with $t=1,2,\cdots,m$, and $C_{n}^{k_{t}}$ is binomial coefficient (i.e., $C_{n}^{k_{t}}=n!/[k_{t}!(n-k_{t})!]$).
\end{lemma}

According to Lemma \ref{lemma.Bernstein}, for a target function $\phi(x)$, we have the following result:
\begin{equation}
\label{eq.Berstein-infty-appro}
\lim_{n\rightarrow + \infty}B_{n}(\phi;x_{1}, \dots, x_{m})=\phi(x_{1}, \dots, x_{m}).
\end{equation}

Then, on the basis of polynomial interpolation theory \cite{Ibikli2008Direct}, we give the following theorem.
\begin{theorem}
\label{theorem.GPS-appro-ability}
For a real-valued function with $m$ variables continuous on $[0,1]^{m\times1}$, it could be approximated by GPS uniformly.
\end{theorem}

\begin{IEEEproof}
Without loss of generality, in light of Eq. (\ref{eq.gegenbauerapproximation}) and  (\ref{eq.error-convergence}), $b_{n,k_{j}}(x_{j})$ can be approximated by GPS as follows:
\begin{equation}
\label{eq.Bernsteinappro}
b_{n,k_{t}}(x_{t})=\sum_{i_{t}=0}^{\infty}\vartheta_{t,i_{t}, k_{t}}g_{i_{t}}^{\lambda}(x_{t}).
\end{equation}
\noindent where $\vartheta_{t,i_{t}, k_{t}}$ denotes the coefficient corresponding to  $g_{i_{t}}^{\lambda}(x_{t})$.

Then, in light of the lemma given above, we could substitute the Eq. (\ref{eq.Bernsteinappro}) into Eq. (\ref{eq.Bernstein}):
\begin{equation}
\begin{aligned}\label{eq.network-basis}
\phi(x)&=\sum^{n}_{k_{1}=0}\cdots\sum^{n}_{k_{m}=0}\phi_{c}\prod_{t=1}^{m}(\sum^{\infty}_{i_{t=0}}\vartheta_{t,i_{t}, k_{t}}g_{i_{t}}^{\lambda}(x_{t}))+\xi_{\phi}(n)\\
&=\sum^{n}_{k_{1}=0}\cdots\sum^{n}_{k_{m}=0}\phi_{c}\prod_{t=1}^{m}(\sum^{n_{t}}_{i_{t=0}}\vartheta_{t,i_{t}, k_{t}}g_{i_{t}}^{\lambda}(x_{t}))+\tilde{\xi}_{\phi}(n))\\
&=\sum^{n_{1}}_{k_{1}=0}\cdots\sum^{n_{m}}_{k_{m}=0}c_{k_{1},\cdots,k_{m}}\prod_{t=1}^{m}g_{k_{t}}^{\lambda}(x_{t})+\tilde{\xi}_{\phi}(n)\\
&=\sum^{L}_{l=1}w_{l}G_{l}^{\lambda}(x)+\tilde{\xi}_{\phi}(n),
\end{aligned}
\end{equation}
\noindent where $n_{t}$ is the total amount of Gegenbauer polynomials utilized to approximate $b_{n,k_{t}}(x_{t})$, $\xi_{\phi}(n)$ and $\tilde{\xi}_{\phi}(n)$ is the residual error functions related to function $\phi(x)$, $c_{k_{1},\cdots,k_{m}}$ is the coefficient for $\prod_{t=1}^{m}g_{k_{t}}^{\lambda}(x_{t})$, $G_{l}^{\lambda}(x)$, termed as the Gegenbauer Elementary function (GEF) in this paper, is the $m$-order tensor product of the one-dimensional Gegenbauer polynomials, which could be expressed as:
\begin{equation}
\label{Bernstein-product}
G_{l}^{\lambda}(x)=G^{\lambda}_{k_{1},\dots, k_{m}}(x)=g_{k_{1}}^{\lambda}(x_{1})\cdots g_{k_{m}}^{\lambda}(x_{m}),
\nonumber
\end{equation}
$w_{l}$ is the weights for $G_{l}^{\lambda}(x)$, and $L=\prod^{m}_{t=1}n_{t}$ is the total amount of Gegenbauer polynomials tensor product (i.e., $\{G_{l}^{\lambda}(x)\}$) utilized to approximate $\phi(x)$.

Besides, according to Eq. (\ref{eq.Bernsteinappro}) and Eq. (\ref{eq.error-convergence}), the approximation residual error function of $\phi(x)$ using GPS would converge to zero:
\begin{equation}
\label{eq.GSP-appro-residual}
\lim_{n,n_{t}\rightarrow +\infty}\tilde{\xi}_{\phi}(n)=0.
\end{equation}
Thus, the proof of GPS for multivariate function approximation is thus completed.
\end{IEEEproof}
Note that the $n, n_{1}, \cdots, n_{t}$ should be large enough to guarantee the approximation capability of the GPS. Therefore, according to Theorem \ref{theorem.GPS-appro-ability}, a series of GEFs  $\{G_{l}^{\lambda}(x)\}$, sorted by graded lexicographer \cite{Stokman2014Orthogonal,Zhang2014Weights}, with optimal weights $w_{l}$ could minimize the residual error function $\tilde{\xi}_{\phi}(n)$ and best approximate the function $\phi(x)$.

\subsection{Network Architecture}
\label{subsec.Network Architecture}
Based on the analysis in Section \ref{subsec.GPS for Multivariate Function Approximation}, the SOGNN and MOGNN are hereby constructed and investigated. The output patterns of both forms are then illustrated.
\subsubsection{SOGNN Model}
\label{Single Output GNN Model}

\begin{figure}[t]

\centering

\includegraphics[width=0.85\columnwidth]{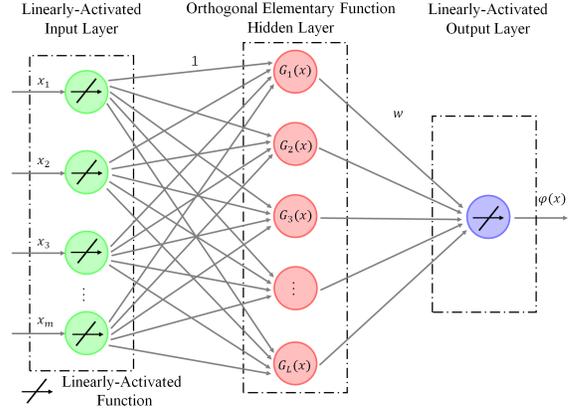}
\caption{The network architecture of GNN.}
\label{fig.SOGNN}
\end{figure}

According to the theoretical analysis above, the architecture of SOGNN is constructed. As is shown in Fig. \ref{fig.SOGNN} , the network architecture of SOGNN consists of three layer, including input layer, hidden layer and output layer.  Let $x=[x_{1},x_{2},\dots,x_{m}]$ denotes the $m$-dimensional input vector, and all of the input data should be normalized to $[0,1]$. In this model, the $m$ neurons in input layer and the single neuron in output layer are both chosen linear in our model. Besides, the activation functions in hidden layer are a series of GEFs $\{G_{l}^{\lambda}(x)| l=1,2,\cdots,L\}$, and $L$ denotes the total amount of hidden neurons. Thus, the output of SOGNN in this case can be expressed as follow:
\begin{equation}
\label{eq.output_of_SOGNN}
\psi(x)=\sum^{L}_{l=1}w_{l}G_{l}^{\lambda}(x)=G(x)w,
\end{equation}
\noindent where $G(x)=[G_{1}^{\lambda}(x), G_{2}^{\lambda}(x),\cdots,G_{L}^{\lambda}(x)]$ denotes the Gegenbauer activation vector with respect to $x$, $G(x)$ actually maps the $m$-dimensional input space to the $L$-dimensional hidden layer feature space, and $w=[w_{1}, w_{2}, \cdots w_{L}]^{\rm T} \in \mathbb{R}^{L}$ denotes the connecting weights vector.

If the number of samples is $S$, their outputs can be written in matrix form, i.e.,
\begin{equation}
\label{eq.output_of_SOGNN}
\Psi(x)=Gw,
\end{equation}
where $\Psi(x) =[\psi(x_1), \psi(x_2),\cdots \psi({x_s})]^{\rm T}$ and
\begin{eqnarray}
\label{eqn.activatematrix_SOGNN}
\begin{aligned}
G &=
\begin{bmatrix}
G(x_{1}) & G(x_{2}) & \cdots  & G(x_{S})
\end{bmatrix}
^{\rm T}\\
&=\begin{bmatrix}
\ G_{1}^{\lambda}(x_{1}) & G_{2}^{\lambda}(x_{1}) & \cdots &G_{L}^{\lambda}(x_{1})\\
 G_{1}^{\lambda}(x_{2}) & G_{2}^{\lambda}(x_{2})  & \cdots &G_{L}^{\lambda}(x_{2})\\
 \vdots & \vdots & \ddots   &  \vdots \\
 G_{1}^{\lambda}(x_{S}) & G_{2}^{\lambda}(x_{S}) & \cdots & G_{L}^{\lambda}(x_{S})
\end{bmatrix}
 \in \mathbb{R}^{S\times L},
\nonumber
\end{aligned}
\end{eqnarray}

Based on Theorem \ref{theorem.GPS-appro-ability} (specifically, Eq. (\ref{eq.network-basis}) and (\ref{eq.GSP-appro-residual}), with $L$ large enough and optimal weights, $\psi(x)$ can best approximate the target function $\phi(x)$. Also, in this model, all the input weights are fixed to be $1$, and the threshold values for all neurons are fixed to be $0$.

Notably, the aforementioned parametric settings can effectively simplify the network architecture, lower the computational complexity, make it more convenient for future implementation on hardware, and have been proved effective in previous works \cite{Zhang2014Weights,JIE2018A}.

Even more, such parametric simplification, on one hand, maintains the approximation capability and the optimality of the network since the network (shown in Fig. \ref{fig.SOGNN}) with the aforementioned parametric settings well matches the theoretical analysis in Eq. (\ref{eq.output_of_SOGNN}) and (\ref{eq.GSP-appro-residual}). On the other hand, such fixed parametric settings could also improve the robustness of the model when faced with disturbance in input. According to \cite{Man2011A,Martin1999Neural}, when the input weights and hidden layer biases are randomly assigned, the robustness could be poor, since the changes of the output weights matrix sometimes could be very large when input disturbance is encountered, which will result in increasing risk of network.


Further, with activation functions being orthogonal elementary functions, our model prevents the hidden activation matrix $G$ from being not full column rank and ill-conditioned, according to orthogonal polynomials theory \cite{Gautschi2004Orthogonal}. Also, with $G^{\rm T}G$ being invertible, the direct computing of $(G^{\rm T}G)^{-1}$ can be easily obtained. Notably, this sets the basis of the computational robustness of the proposed method in Section \ref{subsec.Equality Constrained-Optimization Problems for SOGNN and MOGNN} \cite{Horata2013Robust}.

\subsubsection{MOGNN Model}
\label{Multiple Output GNN Model}

\begin{figure}[t]

\centering

\includegraphics[width=0.85\columnwidth]{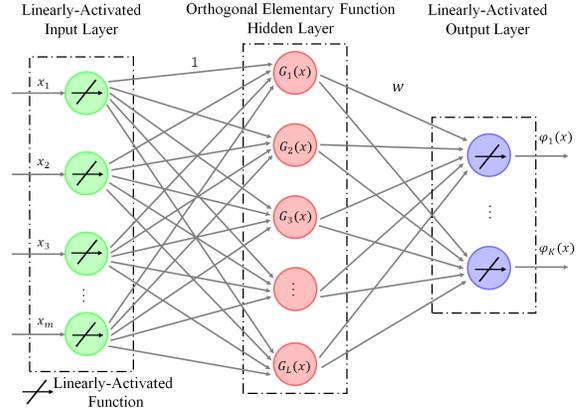}
\caption{The network architecture of GNN.}
\label{fig.MOGNN}
\end{figure}

As is shown in Fig. \ref{fig.MOGNN}, the MOGNN model is constructed based on SOGNN. To be more specific, the MOGNN with $K$ outputs consists of $K$ SOGNN. The output of MOGNN is $K$-dimensional vector, where $\psi(x)=[\psi(x)_{1},\psi(x)_{2},\dots,\psi(x)_{K}]$ and the $k$th output of MOGNN $\psi(x)_{k}$ is the output of the $k$th SOGNN. The parametric settings is the same as the settings of SOGNN. The output of $k$th output node is given below:
\begin{equation}
\label{eq.output_of_GOGNN}
\psi(x)_{k}=\sum^{L_{k}}_{l_{k}=1}w_{l_{k}}G_{l}^{\lambda}(x)=G(x)w,
\end{equation}

\section{Proposed Regularized Weights Direct Determination}
\label{sec.Proposed Regularized Weights Direct Determination}
In this section, the detailed explanation of our proposed R-WDD for GNN is presented. First, the original WDD method is reviewed, after which the main drawbacks of it are discussed. Then, based on equality-constrained optimization, a method minimizing the empirical risks and structural risks with better generalization ability is provided, and the analysis of it is carried out in cases of two forms of GNN, respectively. Finally, the unified R-WDD for GNN weights determination under different cases are given.

\subsection{Review of the Original Weights Direct Determination}
\label{subsec.Review of the Original Weights Direct Determination}
According to the previous works in \cite{JIE2018A,Zhang2014Weights}, for a given GNN model, the objective loss function of WDD is:

\begin{eqnarray}
\begin{split}
\text{minimize:~~}  &\parallel\Psi-\Phi\parallel^{2}_{2}=\parallel Gw-\Phi\parallel^{2}_{2}.
\end{split}
\label{eqn.WDD_objective}
\end{eqnarray}
where $\Phi =[\phi(x_1), \phi(x_2),\cdots \phi({x_s})]^{\rm T}$ is the label of $s$ samples.

With WDD, the solution of the optimal output weights can be calculated by the following equation:
\begin{equation}
\label{eq.original_WDD}
w=(G^{\rm T}G)^{-1}G^{\rm T}\Phi=G^{\rm \dag}\Phi,
\end{equation}
\noindent where $(G^{\rm T}G)^{-1}G^{\rm T}$ is the Moore-Penrose generalized inverse of activation matrix $G$, also termed as $G^{\dag}$.

According to \cite{Banerjee1971Generalized}, there are several methods to calculate the Moore-Penrose generalized inverse of a matrix: orthogonal projection method, iterative method and singular values decomposition (SVD).


Since the WDD directly calculates the least-square solution involving an inversion of a matrix, the speed of calculation can be guaranteed. Further, according to the previous works \cite{JIE2018A,Zhang2014Weights}, the networks equipped with WDD have demonstrated outstanding performances, such as excellent approximation ability, easy implementation.

However, although WDD method has aforementioned merits, it still has a main drawback. As is shown Eq. (\ref{eqn.WDD_objective}), the WDD only focuses on empirical risk minimization and no measure is taken to prevent the model hypothesis from being too complicated, which might easily lead to an over-fitting model \cite{Vapnik1995The,Bartlett1998The}.
\subsection{Equality-constrained Optimization Problems for GNN}
\label{subsec.Equality Constrained-Optimization Problems for SOGNN and MOGNN}
Due to the aforementioned drawbacks of the original WDD method, in this paper, our model does not only tend to seek the smallest training error but also seek the smallest norm of output weights. According to the famous Barlett's theory \cite{Bartlett1998The}, a good model with generalization ability should be the one could reach a best trade-off between the empirical risk and the structural risk. Generally, the empirical risk is the training error of the neural network, which could be represented by $\ell_{2}$-norm, i.e., $\parallel\xi\parallel^{2}_{2}=\parallel \Psi-\Phi\parallel^{2}_{2}$; the structural risk is the $\ell_{2}$-norm of output weights, i.e., $\parallel w\parallel_{2}$, which denotes the complexity of the model\cite{Suykens1999Least}. Thus, the optimization object and the mathematical model could be formulated as below:

\begin{eqnarray}
\begin{split}
\text{minimize:~~}  &\parallel Gw-\phi\parallel_{2}^{2} \text{and} \parallel w\parallel_{2}.
\end{split}
\label{eqn.optimization_objective_L2}
\end{eqnarray}


According to the ridge regression theory \cite{Hoerl2000Ridge}, it would make solution much stabler and tending to have better generalization ability since a positive value added to the diagonal of $(G^{\rm T}G)^{-1}G^{\rm T}$ or $G^{\rm T}(GG^{\rm T})^{-1})$. Also, according to the statistical learning theory \cite{Martin1999Neural,Suykens2002Weighted}, the real risk of a model should be the sum of the empirical one and the structural one, and these two kinds of risk should be considered together in discussions. Since the proposed GNN has two forms (i.e., SOGNN and MOGNN), in the following, the solution for each form would be discussed respectively. After that, it can be seen the solution for SOGNN is actually one specific case of MOGNN. Thus, these two solutions are finally unified, which leads to the final solution of R-WDD.

\subsubsection{SOGNN for Classification}
Based on the analysis in Section \ref{sec.Theoretical Basis for GNN}, the SOGNN can approximate any continuous target function in the variables space. For classification problems, the expected output of SOGNN can be close enough to the target class labels in the corresponding regions. Thus, the objective function of proposed R-WDD for SOGNN can be described as below:

\begin{eqnarray}
\begin{split}
\text{minimize:~~~}  &\frac{\gamma}{2}\parallel w\parallel_{2}^{2}+\frac{1}{2}\sum^{S}_{s=1}\xi^{2}_{s}\\
\text{subject to:} &\sum^{L}_{l=1}w_{l}G_{l}^{\lambda}(x_{s})-\phi_{s}-\xi_{s}=0, s=1,2,\dots,S,
\end{split}
\label{eqn.Proposed Equality Constrained-Optimization_SOGNN}
\end{eqnarray}
\noindent where the weight factor $\gamma$ is the regularizer. By tuning the values of $\gamma$, the proportion of the empirical risk and the structural risk can be adjusted for reaching the best trade-off. Based on KKT theorem \cite{Tapia1980Practical}, to solve the R-WDD optimization problem for SOGNN is equivalent to solve the following Lagrangian equation:
\begin{eqnarray}
\begin{split}
\mathcal{L}\big(w,\xi,\alpha)&=\frac{\gamma}{2}\parallel w\parallel^{2}_{2}+\frac{1}{2}\sum^{S}_{s=1}\xi^{2}(x_j)\\
&-\sum^{S}_{s=1}\alpha_{s}(G(x_{s})w-\phi(x_s)+\xi(x_s))\\
&=\frac{\gamma}{2}\parallel w\parallel^{2}_{2}+\frac{1}{2}\parallel \xi\parallel^{2}_{2}-\alpha(Gw-\Phi+\xi),
\label{eqn.Lagrange_SOGNN}
\end{split}
\end{eqnarray}
where $\alpha\in \mathbb{R}^s$ denotes the Lagrangian multiplier vector with equality constraints of Eq. (\ref{eqn.Proposed Equality Constrained-Optimization_SOGNN}) and $\alpha=[\alpha_{1},\dots,\alpha_{S}]$. By setting the gradients of the Lagrangian equation with respect to $w$, $\xi$ and $\alpha$ equal to zero respectively, we can obtain the the following KKT optimality conditions of Eq. (\ref{eqn.Lagrange_SOGNN}):
\begin{subequations}
\begin{eqnarray}
\begin{split}
&\frac{\partial \mathcal{L}\big(w,\xi,\alpha)}{\partial w}=0 &~~~~\rightarrow~~&\gamma w- G^{\rm T}\alpha=0,
\label{eqn.Lagrange_SOGNN-2-1}
\end{split}
\end{eqnarray}
\begin{eqnarray}
\begin{split}
&\frac{\partial \mathcal{L}\big(w,\xi,\alpha)}{\partial \xi}=0~~~~&\rightarrow~~~&\xi-\alpha=0,~~~
\label{eqn.Lagrange_SOGNN-2-2}
\end{split}
\end{eqnarray}
\begin{eqnarray}
\begin{split}
&\frac{\partial \mathcal{L}\big(w,\xi,\alpha)}{\partial \alpha}=0&~~~~\rightarrow~~&Gw-\Phi+\xi=0.
\label{eqn.Lagrange_SOGNN-2-3}
\end{split}
\end{eqnarray}
\label{eqn.Lagrange_SOGNN-2}
\end{subequations}

\subsubsection{MOGNN for Classification}
According to Section \ref{subsec.Network Architecture}, the MOGNN is actually a generalized form of the SOGNN. For $K$-classification problems, the number of the output nodes of MOGNN should be $K$, and when the original class label of sample is $k$, the expected output of MOGNN should be a $K$ dimensional vector and the $k$th component of it should be close to $1$, while others should be close to $-1$. Thus, the objective function of proposed R-WDD for MOGNN can be described as below:

\begin{eqnarray}
\begin{split}
\text{minimize:~~}  &\frac{\gamma}{2}\parallel w\parallel_{\rm F}^{2}+\frac{1}{2}\sum^{S}_{s=1}\parallel\xi_{s}\parallel_{2}^{2}\\
\text{subject to:} &G(x_{s})w-\phi_{s}+\xi_{s}=0, s=1,2,\dots,S,
\end{split}
\label{eqn.Proposed Equality Constrained-Optimization_MOGNN}
\end{eqnarray}
\noindent where $\phi_{s}=[\phi_{s,1},\phi_{s,2},\dots,\phi_{s,K}]$ is the target vector corresponding to sample $s$, $\xi_{s}=[\xi_{s,1},\xi_{s,2},\dots,\xi_{s,K}]$ is the error vector, $w$ is a matrix of $L\times K$, where $w_{k}$ is the vector of the output weights corresponding to $k$th output node and $\parallel \cdot \parallel_{\rm F}$ is Frobenius norm.

Similar to SOGNN, the Lagrangian equation for R-WDD optimization for MOGNN is:
\begin{eqnarray}
\begin{split}
\mathcal{L}\big(w,\xi,\alpha)&=\frac{\gamma}{2}\parallel w\parallel^{2}_{F}+\frac{1}{2}\sum^{S}_{s=1}\parallel\xi_{s}\parallel^{2}_{s}\\
&-\sum^{K}_{k=1}\sum^{S}_{s=1}\alpha_{k,s}(G(x_{s})w-\phi_{s,k}+\xi_{s,k})\\
&=\frac{\gamma}{2}\parallel w\parallel^{2}_{F}+\frac{1}{2}\parallel \xi\parallel^{2}_{F}-\alpha(GW-\phi+\xi),
\label{eqn.Lagrange_MOGNN}
\end{split}
\end{eqnarray}
\noindent where $\alpha\in \mathbb{R}^{K\times S}$ denotes the Lagrangian multiplier matrix, $\alpha_{s}=[\alpha_{s,1},\dots,\alpha_{s,K}]^{\rm T}$ and $\alpha=[\alpha_{1},\dots,\alpha_{S}]$. By setting the partial derivatives of Eq. (\ref{eqn.Lagrange_MOGNN}) equal to zero respectively, we can also obtain the KKT optimality conditions of Eq. (\ref{eqn.Lagrange_MOGNN}) as follows:
\begin{subequations}
\begin{eqnarray}
\begin{split}
&\frac{\partial \mathcal{L}\big(w,\xi,\alpha)}{\partial w}=0 &~~~\rightarrow~~~&\gamma w- G^{\rm T}\alpha=0,
\label{eqn.Lagrange_MOGNN-2-1}
\end{split}
\end{eqnarray}
\begin{eqnarray}
\begin{split}
&\frac{\partial \mathcal{L}\big(w,\xi,\alpha)}{\partial \xi}=0~~~~&\rightarrow~~~~~&\xi-\alpha=0,~~~
\label{eqn.Lagrange_MOGNN-2-2}
\end{split}
\end{eqnarray}
\begin{eqnarray}
\begin{split}
&\frac{\partial \mathcal{L}\big(w,\xi,\alpha)}{\partial \alpha}=0&~~~\rightarrow~~&Gw-\Phi+\xi=0.
\label{eqn.Lagrange_MOGNN-2-3}
\end{split}
\end{eqnarray}
\label{eqn.Lagrange_MOGNN-2}
\end{subequations}

It can be seen from the Eq. (\ref{eqn.Lagrange_SOGNN-2}) and Eq. (\ref{eqn.Lagrange_MOGNN-2}) that to solve the R-WDD optimization for SOGNN is actually to solve one particular case of the R-WDD optimization for MOGNN when the number of output nodes $k$ is fixed to $1$. To be more specific, some of the matrices relevant to $K$ in MOGNN reduce to vectors in SOGNN. Thus, the solutions of weights for SOGNN and MOGNN could be analyzed together. It is worth mentioning that the size of the activation matrix $G$ remains the same in the both cases, and it only depends on the number of training samples $S$ and the neuron number $L$.

\subsection{Proposed Regularized Weights Direct Determination}
\label{subsec.Proposed Regularized Weights Direct Determination}
Based on the aforementioned KKT optimality conditions for SOGNN and MOGNN, the unified solution of output weights for SOGNN and MOGNN can be obtained. However, due to the efficiency concerns, different solutions to the above KKT optimality conditions can be obtained based on the size of the training data sets.

\subsubsection{Small Training Data Sets}
When $S$ is not huge, or more specifically, less than $L$, by substituting Eq. (\ref{eqn.Lagrange_SOGNN-2-1}) and (\ref{eqn.Lagrange_MOGNN-2-1}), and Eq. (\ref{eqn.Lagrange_SOGNN-2-2}) and (\ref{eqn.Lagrange_MOGNN-2-2}) into Eq. (\ref{eqn.Lagrange_SOGNN-2-3}) and (\ref{eqn.Lagrange_MOGNN-2-3}) respectively, we can get:
\begin{equation}
\label{eq.weights_solution1_phi}
(\gamma I+GG^{\rm T})\alpha = \Phi,
\end{equation}
where $I$ is an identity matrix, whose size is $S\times S$, and the same as the size of $GG^{\rm T}$.

By substituting Eq.(\ref{eq.weights_solution1_phi}) into Eq. (\ref{eqn.Lagrange_SOGNN-2-1}) and (\ref{eqn.Lagrange_MOGNN-2-1}), and getting rid of $\alpha$, the following equation can be obtained:

\begin{equation}
\label{eq.weights_solution1}
w = G^{\rm T}(\gamma I+GG^{\rm T})\Phi,
\end{equation}
when the number of training samples is not large, the calculation of R-WDD mainly involves an inversion of a $S\times S$ matrix. It is worth mentioning: for SOGNN, the $w$ and $\Phi$ is column vectors; for MOGNN, the $w$ and $\Phi$ evolve to matrices.

\subsubsection{Huge Training Data Sets}
When $S$ is huge, i.e., greater than $L$, by substituting Eq. (\ref{eqn.Lagrange_SOGNN-2-3}) and (\ref{eqn.Lagrange_MOGNN-2-3}) into Eq. (\ref{eqn.Lagrange_SOGNN-2-2}) and (\ref{eqn.Lagrange_MOGNN-2-2}) respectively, the explicit expression for $\alpha$ can be obtained as follows:
\begin{equation}
\label{eq.alpha-1}
\alpha=-(Gw-\Phi).
\end{equation}

Based on this, we can obtain the following solution for weights vector $w$, by substituting Eq. (\ref{eq.alpha-1}) into Eq. (\ref{eqn.Lagrange_MOGNN-2-1}):
\begin{equation}
\label{eq.weights_solution}
w=(\gamma I+G^{\rm T}G)^{-1}G^{\rm T}\Phi,
\end{equation}
where $I$ is an $L\times L$ identity matrix, the same as the size of $G^{\rm T}G$. Thus, the calculation mainly involves an inversion of an $L\times L$ matrix, and the speed only hinges on $L$ when $S$ grows large.

Based on the aforementioned discussion under two circumstances and Eq. (\ref{eq.output_of_SOGNN}), the final network output equation is:

\begin{equation}
\label{eq.network_output_equation}
\Psi(x)=\left\{
\begin{array}{ll}
G(x)w=G^{\rm T}(\gamma I+GG^{\rm T})\Phi, & S \leq L\\
G(x)w=(\gamma I+G^{\rm T}G)^{-1}G^{\rm T}\Phi, & S > L
\end{array}
\right.
\end{equation}

As is shown in the Eq. (\ref{eq.network_output_equation}), the calculation of R-WDD always only involves an inversion of a matrix of relatively small size, thus the the rapidness of R-WDD calculation is guaranteed.

\begin{table*}[]

\scriptsize

\centering

\caption{Specifications of Experiments of Binary Classification Problems}

\label{Tab:Specification of Experiments of Binary Classification Problems}

\begin{tabular}{ccccccccccccc}

\toprule

\multirow{2}{*}{Data Sets}&\multirow{2}{*}{Training}&\multirow{2}{*}{Testing}&\multirow{2}{*}{Classes} &\multirow{2}{*}{Features}& \multicolumn{2}{c}{LS-SVM} & \multicolumn{3}{c}{ELM} & \multicolumn{3}{c}{GNN} \\

\cmidrule(r){6-7} \cmidrule(r){8-10} \cmidrule(r){11-13}

&&&&&  $C$     &  $\delta $

&  $C$       &  $\delta$   &   $L$

&  $\gamma$     &   $L$ &   $\lambda$\\

\midrule

Australian  & 484& 206  &2&6     &    $2^{7}$      &      $2^{10}$              & $2^{1}$      & $2^{4}$     & 1500   &  $2^{-12}$  & 1500                        &   0.05\\

Banana  & 1591& 3709   &2&  2  &    $2^{14}$      &      $2^{2}$              & $2^{0}$      & $2^{4}$     & 1500   &   $2^{-12}$ & 1500    &   0.05\\

Diabetes & 537& 231   &2&8    &    $2^{10}$      &      $2^{10}$              & $2^{10}$      & $2^{6}$     & 1500   &  $2^{-8}$  & 1500                        &  0.05\\

Liver & 241& 104   &2&   6 &    $2^{5}$      &      $2^{7}$              & $2^{8}$      & $2^{5}$     & 1500   &  $2^{-13}$  & 1500                        &   0.05\\

Ionosphere  & 245& 106 &2&  33    &    $2^{5}$      &      $2^{7}$              & $2^{5}$      & $2^{6}$     & 1500   &  $2^{-11}$  & 1500                        &   0.05\\
\bottomrule

\end{tabular}

\end{table*}

\begin{table*}[]

\scriptsize

\centering

\caption{Specifications of Experiments of Multi-Classification Problems}

\label{Tab:Specification of Experiments of Multiclassification Classification Problems}

\begin{tabular}{ccccccccccccc}

\toprule

\multirow{2}{*}{Data Sets}&\multirow{2}{*}{Training}&\multirow{2}{*}{Testing}&\multirow{2}{*}{Classes} &\multirow{2}{*}{Features} & \multicolumn{2}{c}{LS-SVM} & \multicolumn{3}{c}{ELM} & \multicolumn{3}{c}{GNN} \\

\cmidrule(r){6-7} \cmidrule(r){8-10} \cmidrule(r){11-13}

&&&&&  $C$     &  $\delta $

&  $C$       &  $\delta$   &   $L$

&  $\gamma$     &   $L$ &   $\lambda$\\

\midrule

Iris  & 105& 45  &3 &  3   &    $2^{14}$      &      $2^{7}$              & $2^{1}$      & $2^{2}$     & 1500   &  $2{-12}$  & 1000                        &   0.05\\

Glass  & 158 & 56 &6&  9 &  $2^{2}$      &      $2^{3}$   & $2^{10}$      & $2^{9}$     & 1500   &   $2^{-12}$ & 1000                       &   0.05\\

Wine& 126& 52    &3&  13 &    $2^{15}$      &      $2^{5}$              & $2^{1}$      & $2^{9}$     & 1500   & $2^{-6}$   & 1000                        &   0.05\\

Ecoli  & 243& 93  & 8 &7&  $2^{7}$      &      $2^{5}$    & $2^{20}$      & $2^{11}$     & 1500   &   $2^{-8}$ & 1000       &  0.05\\

Vehicle & 594& 252   &4&  18  &    $2^{24}$      &      $2^{11}$     & $2^{6}$      & $2^{6}$     & 1500   &  $2^{-14}$  & 1000                        &   0.05\\

\bottomrule

\end{tabular}

\end{table*}
\section{Experiments and Analysis}
\label{sec.experiments and Analysis}
In this section, to verify the performance of the GNN with proposed R-WDD, we first compare the generalization performance of GNN with different algorithms (i.e., LS-SVM with Gaussian kernel and ELM with Gaussian kernel, another two famous algorithm based on equality-constrained optimization) on real-world benchmark binary, multi-class classification cases. To make it more intuitive, the decision boundaries of different algorithms are presented to demonstrate the classification capability of the proposed method. Then, we also compare the computational scalability of three algorithms, and the computational efficiency of GNN and ELM. Since model's sensitivity to hyper parameter selection is also one key feature of algorithms, experiments between GNN and ELM are conducted to test model's tolerance for the hyper parametric changes. At last, we conduct an experiment in which different amplitude noises are added to test the robustness to random noises of the different algorithms.

The simulations in Subsection \ref{subsec.Performance Comparison on Real-World Benchmark Data Sets} and \ref{subsec.Comparison Robustness of Classification} are carried out in MATLAB R2018b environment running in Intel(R) Core(TM) i5-8520U 1.60 GHz CPU with 8-GB RAM, and the simulations in Subsection \ref{subsec.Computational Scalability and Efficiency} and \ref{subsec.Sensitivity to Hyper Parameters} are carried out in MATLAB R2018a environment running in Intel(R) Core(TM) i7-8700T 2.40GHz CPU with 16-GB RAM.The codes for LS-SVM and ELM are downloaded from \cite{LS-SVM} and \cite{ELM-code} respectively.
\subsection{Benchmark Data Sets and Hyper Parameters Selections}
\label{subsec.Benchmark Data Sets and Hyper Parameters Selections_5}
To verify the classification performance of different algorithms, 5 binary classification cases and 5 multi-classification cases have been tested in out experiments. All data sets are downloaded from the UCI Machine Learning Repository \cite{UCIdatasets}.

According to \cite{Huang2012Extreme}, the selection of hyper parameters greatly influences the performance of the algorithms. For LS-SVM with the popular Gaussian kernel (i.e., $K(u,v)={\rm exp}(-\delta||u-v||^{2})$), it mainly has two hyper parameters which are the cost parameter $C$ (preventing the model from  being too complex) and kernel parameter $\delta$. ELM using Gaussian kernel, has three hyper parameters, the cost parameter $C$, kernel parameter $\delta$ and number of nodes $L$. For the proposed GNN with R-WDD, it also has three hyper parameters which are the regularizer $\gamma$, the parameter for GPS $\lambda$ and the number of neurons $L$. In this paper, the value of $\lambda$ is fixed to $0.05$. Also, the number of nodes for ELM and the proposed GNN with R-WDD for binary and multi-classification problems are fixed to 1000.

Specifically, for a binary problem, only one LS-SVM, one ELM with single output node and one SOGNN are trained; for a $K$-class multi-classification problem, it need to train $K$ binary LS-SVM, while for ELM and GNN, only one network with $K$ output nodes is required. Thus, ELM and GNN are more flexible and applicable mechanisms in multi-classification.

In experiments, for LS-SVM and ELM, the hyper parameters (i.e., $\delta$ and $C$) are searched from 31 different values in the fashion of 4-fold cross-validation gird search, which will result in 961 different combinations of ($\delta$,$C$). And the ranges of values of $\delta$ and $C$ are $\{2^{-30},2^{-28},\dots,2^{28},2^{30}\}$. The value of $\gamma$ for GNN is also searched from 31 different values in $\{2^{-30},2^{-28},\dots,2^{28},2^{30}\}$. At last, the specifications of each experiment (including the divisions of data sets, the number of features and classes of data set, and the hyper parameters selections for each algorithm) are reported in TABLE \ref{Tab:Specification of Experiments of Binary Classification Problems} and \ref{Tab:Specification of Experiments of Multiclassification Classification Problems}.

Note that for each experiment, fifty trials are conducted to ensure the results are not accidental values, and the average result is reported. Besides, all the data are reshuffled at each trial and all the data are normalized before experiments.

\subsection{Performance Comparison on Real-World Benchmark Data Sets}
\label{subsec.Performance Comparison on Real-World Benchmark Data Sets}
TABLE \ref{Tab:Performance Comparison of LS-SVM, ELM and MOGNN: Binary Classification Data Sets} -\ref{Tab:Performance Comparison of LS-SVM, ELM and MOGNN: Multiclassification Data Sets} present the performance comparisons of LS-SVM with Gaussian kernel, ELM with Gaussian kernel and GNN with R-WDD. Note that the simulation results include the mean testing classification accuracies (Acc), corresponding mean standard deviations (Std) and the average training time, and the best results are highlighted in boldface.

For binary classification problems, GNN could achieve comparable or even better generalization performance as LS-SVM and ELM with comparably fast learning speed. Take the results of Banana data set as instance.
\begin{itemize}
  \item For Banana data set, the best Acc is achieved by LS-SVM, while the Acc by GNN is only 0.11\% less. Also, GNN demonstrates comparably extremely fast training speed as ELM, with difference within 0.01s.
\end{itemize}

For multi-classification problems, the solutions achieved by GNN are better than LS-SVM and ELM, with higher Acc, and stabler with lower Std. To be illustrative, take the results of Glass and Ecoli data sets as examples.
\begin{itemize}
  \item For Glass data set, Acc of GNN is much higher, with 5.46\% and 4.47\% higher than the results achieved by LS-SVM and ELM.
  \item For Ecoli data set, Acc of the GNN is at least 2\% higher than LS-SVM and ELM, while Std is much lower. Although the training time is a little longer, the training speed of GNN with R-WDD is very fast and accessible.
\end{itemize}
\begin{table*}[]

\scriptsize

\centering

\caption{Performance Comparison of LS-SVM, ELM and GNN: Binary Class Data Sets}

\label{Tab:Performance Comparison of LS-SVM, ELM and MOGNN: Binary Classification Data Sets}

\begin{tabular}{cccccccccc}

\toprule

\multirow{2}{*}{Data Sets} & \multicolumn{3}{c}{LS-SVM} & \multicolumn{3}{c}{ELM} & \multicolumn{3}{c}{GNN} \\

\cmidrule(r){2-4} \cmidrule(r){5-7} \cmidrule(r){8-10}

&  Testing      &  Testing   &   Training

&  Testing       &  Testing   &   Training

&  Testing       &  Testing    &   Training \\

&  Acc (\%)      &  Std (\%)  &   Time (s)

&  Acc (\%)      &  Std (\%)   &   Time (s)

&  Acc (\%)      &  Std (\%)   &   Time (s) \\

\midrule

Australian &    72.08     &    3.44& 0.2462      & 74.69        & 1.42  & 0.2267         &   \textbf{75.91}                         &    3.11          & 0.2639   \\

Banana &\textbf{89.63}   & 0.33    & 0.1241        &89.04    & 0.46  & 0.1124    & 89.52          & 0.61         & 0.1206          \\

Diabetes  &77.01   & 2.41         & 0.1550     & \textbf{77.52}    & 2.46     & 0.1328          & 77.29     & 2.05    & 0.1421          \\

Liver &    71.17  &  3.92   & 0.2541      &\textbf{74.87}     & 1.44     & 0.2817    &  72.68   &  4.04  & 0.2370      \\

Ionosphere &91.58   & 2.74  & 0.2151       & \textbf{92.12}      & 4.73   & 0.1761          & 91.03   & 2.66    & 0.2240          \\

\bottomrule

\end{tabular}

\end{table*}
\begin{table*}[]

\scriptsize

\centering

\caption{Performance Comparison of LS-SVM, ELM and GNN: Multi-Class Data Sets}

\label{Tab:Performance Comparison of LS-SVM, ELM and MOGNN: Multiclassification Data Sets}

\begin{tabular}{cccccccccc}

\toprule

\multirow{2}{*}{Data Sets} & \multicolumn{3}{c}{LS-SVM} & \multicolumn{3}{c}{ELM} & \multicolumn{3}{c}{GNN} \\

\cmidrule(r){2-4} \cmidrule(r){5-7} \cmidrule(r){8-10}

&  Testing      &  Testing   &   Training

&  Testing       &  Testing   &   Training

&  Testing       &  Testing    &   Training \\

&  Acc (\%)      &  Std (\%)  &   Time (s)

&  Acc (\%)      &  Std (\%)   &   Time (s)

&  Acc (\%)      &  Std (\%)   &   Time (s) \\

\midrule

Iris   &    96.22  &      2.36    & 0.0231      & 96.04    & 2.37    & 0.0222  &   \textbf{96.58}      &      2.34   & 0.0257   \\

Glass     &67.32   & 5.04    & 0.0417    & 68.41           & 4.27    & 0.0326   & \textbf{72.68}   & 4.94  & 0.0421   \\

Wine&97.63          & 1.82   & 0.0343       & 98.48           & 4.46           & 0.0326  & \textbf{98.54}      & 1.63       & 0.0457  \\

Ecoli &85.93   & 3.52    & 0.0344    & 87.48        & 2.91      & 0.0382     & \textbf{89.55} & 2.23          & 0.0481   \\

Vehicle  &83.19   & 1.93    & 0.1029     & 83.16        & 1.89      & 0.0931  & \textbf{84.05} & 1.75 & 0.1266     \\
\bottomrule

\end{tabular}

\end{table*}

\subsection{Comparison of Decision Boundaries}
\begin{figure*}[t]

\centering
  \psfrag{X1}[c][c][0.85]{$x_1$}
  \psfrag{X2}[c][c][0.85]{$x_2$}

\subfigure[LS-SVM with $C=2^{-5}$, $\delta=2^{0}$]{\includegraphics[width=0.65\columnwidth]{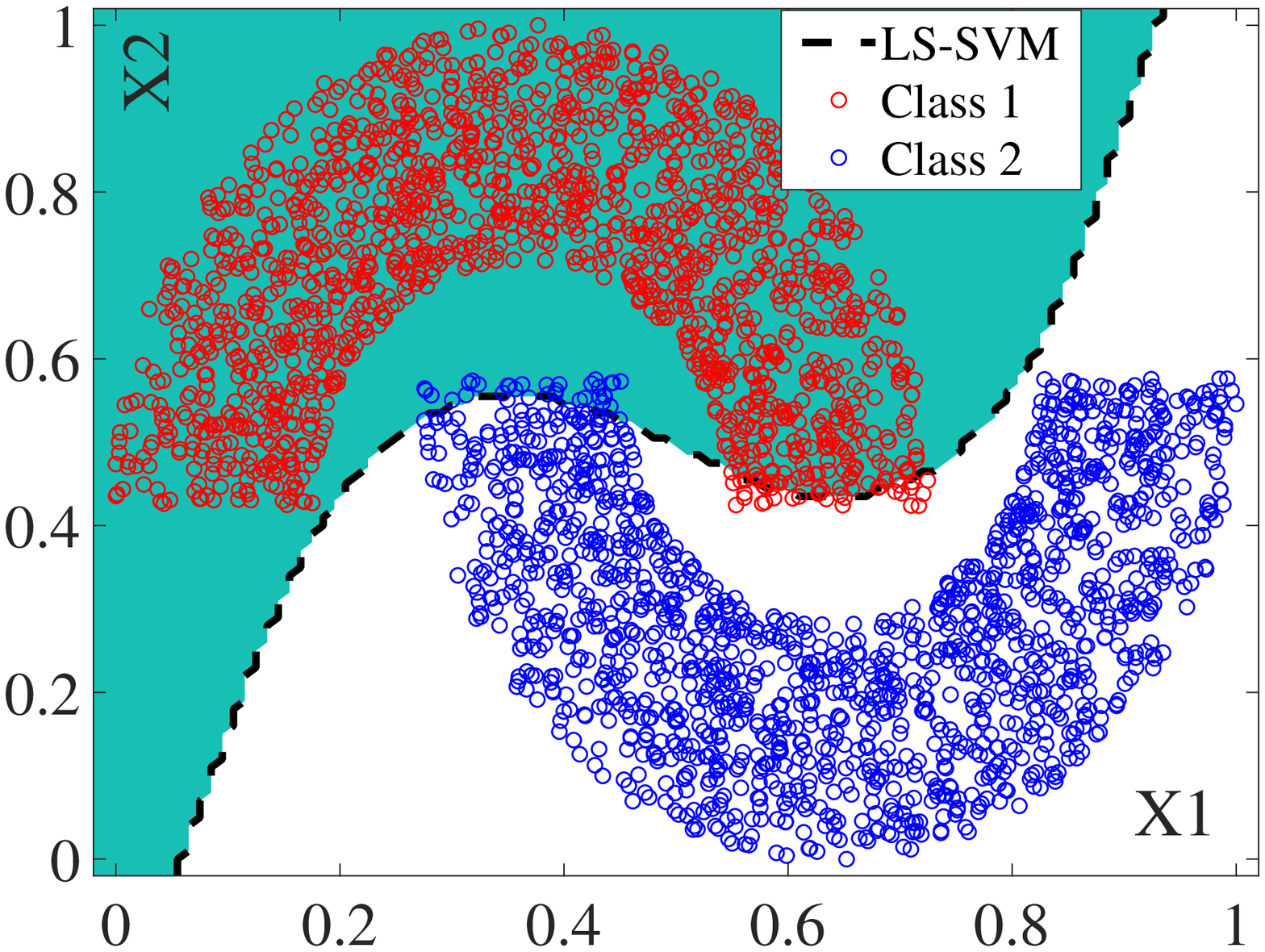}}
\subfigure[ELM with $C=2^{-2}$, $\delta=2^{4}$, $L=100$]{\includegraphics[width=0.65\columnwidth]{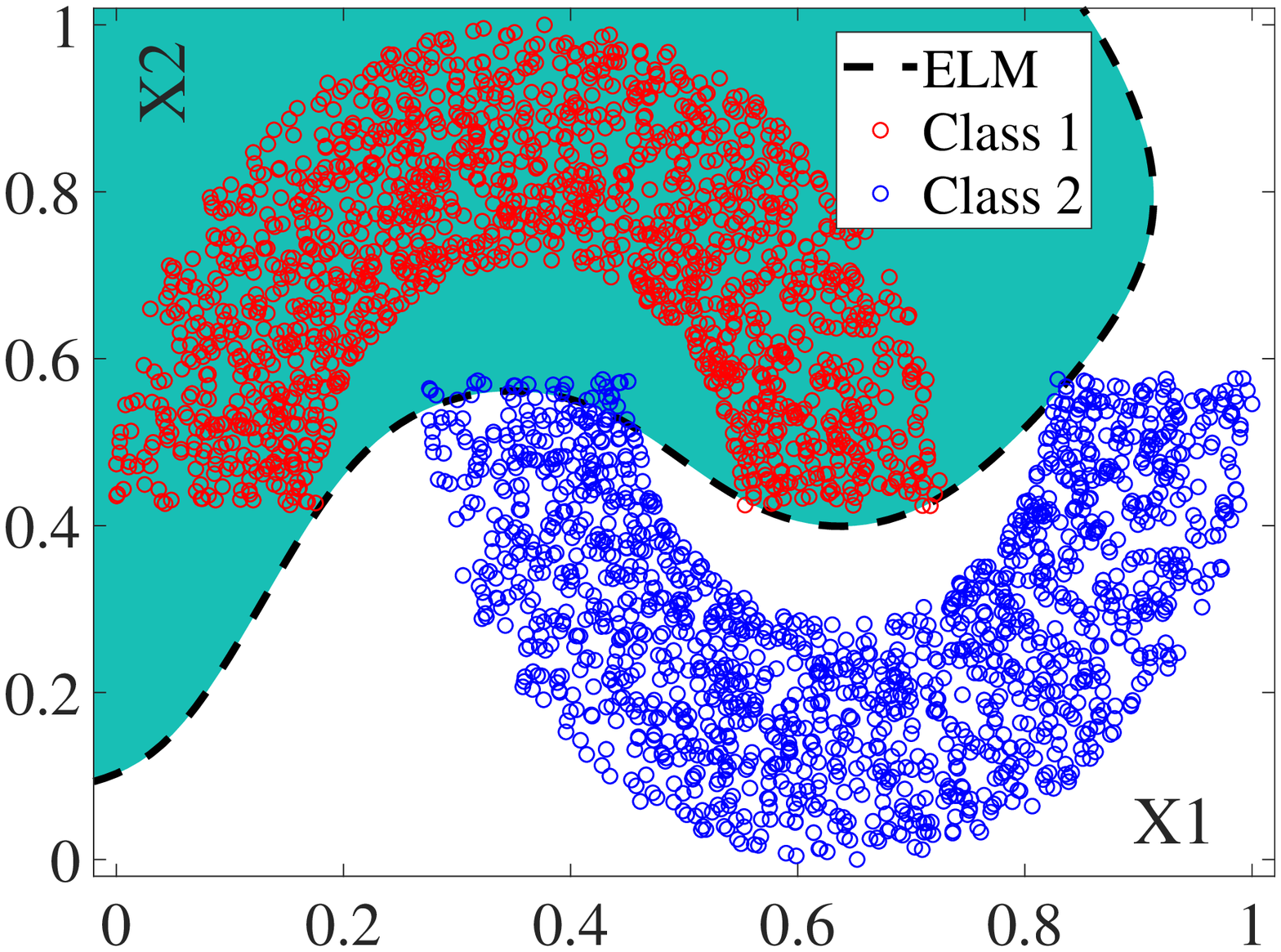}}
\subfigure[GNN with $\gamma=2^{-6}$, $L=100$]{\includegraphics[width=0.65\columnwidth]{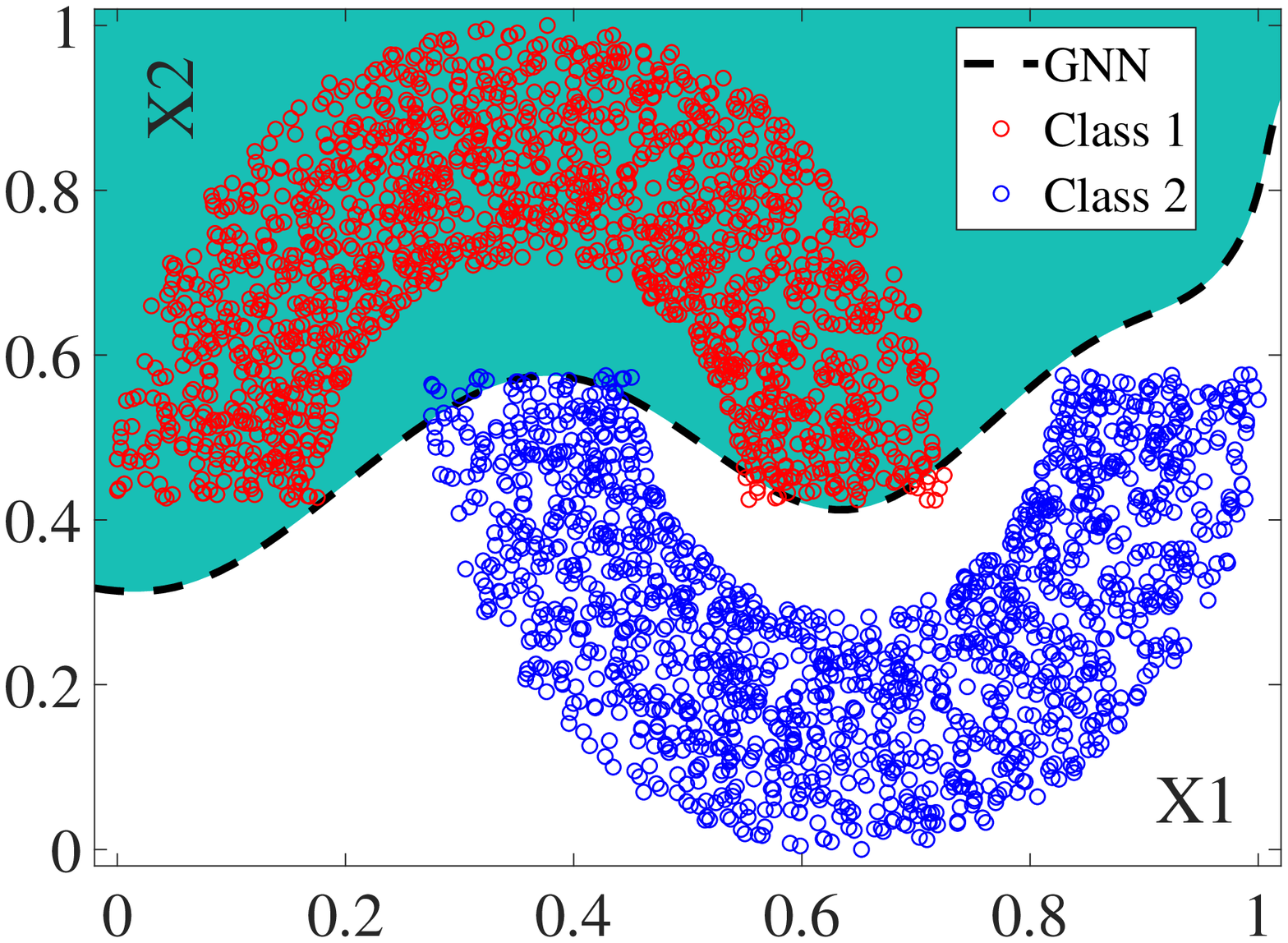}}
\caption{The decision boundaries of each algorithm. It can be observed that three algorithms can classify the nonlinear separable data set well.}
\label{fig.decision_curve}
\end{figure*}

As a more intuitive proof of GNN’s classification capability,we also conduct a simple experiment aiming at showing the decision boundaries of different algorithms on a nonlinear separable data classes. For this experiment, it is conducted on a synthetic two-dimensional, two class data set called double moon, with a pattern of two moons intertwining. Since it is a binary classification problem,  the pattern vectors of Class 1 is tagged ``+1'' labels and ones of Class 2 received ``-1'' labels.

For this experiment, all the classifiers are trained with a data set with 3000 samples. The decision boundary of each algorithm is recorded in Fig. \ref{fig.decision_curve} (a)-(c), respectively. Note that this simple experiment is set to demonstrate the classification ability of each classifier intuitively, thus all the samples were used to train and the testing accuracy rate were not given.

\subsection{Computational Scalability and Efficiency}
\label{subsec.Computational Scalability and Efficiency}

One of the biggest merits of ELM originates from its excellent computational scalability and efficiency, which means the training speed of ELM maintains when the scale of problem grows and the number of hidden neurons $L$ increases. In this subsection, experiments on Liver data sets are conducted to verify the computational scalability and efficiency of GNN. To be specific, since the experiments here is set to evaluate the computational scalability and efficiency of algorithms, we simply amplify size of the Liver data set by randomly copying the samples. To avoid accidental results, each data point plotted is the average values of 100 trials.
\begin{figure}[]

\centering
  \psfrag{S}[c][c][0.85]{$S$}

\includegraphics[width=0.75\columnwidth]{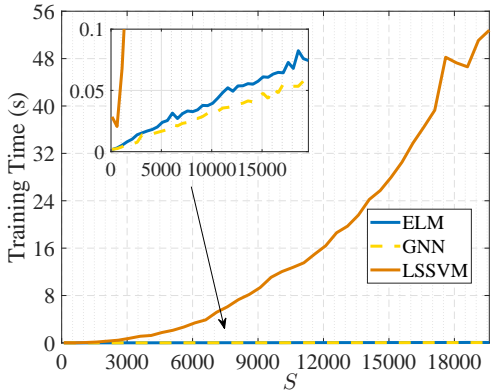}
\caption{Computational scalability of different algorithms: this experiment runs on the amplified Liver data set (20000 samples), and the number of neurons of GNN and ELM is 100. We can see that as $S$ grows, the training time consumed by LS-SVM exponentially increases, while the training time consumed by ELM (Gaussian kernel) and GNN increases very slowly.
}
\label{fig.Samples_Time_exp}
\end{figure}

\subsubsection{Computational Scalability} In the simple example shown in Fig. (\ref{fig.Samples_Time_exp}), we can see the GNN has comparable (or even better) computational scalability as ELM. Specifically, according to the analysis in \cite{Huang2012Extreme} , for LS-SVM the main computational cost comes from its calculating Lagrange multipliers $\alpha$, which is highly associated with the size of data sets $S$. However, since ELM and GNN with R-WDD have different equations calculating the output weights when $S$ is great (refer Eq. (\ref{eq.network_output_equation})), the computational cost of ELM and GNN reduces dramatically. Thus, ELM and GNN has better computational scalability, compared to LS-SVM.

\subsubsection{Computational Efficiency} 
\begin{figure}[t]

\centering
  \psfrag{L}[c][c][0.85]{$L$}

\includegraphics[width=0.75\columnwidth]{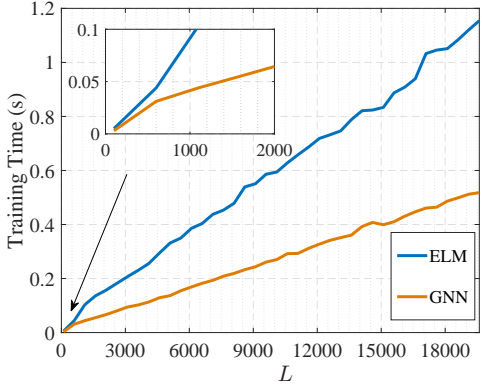}
\caption{Computational efficiency of ELM and GNN: this experiment runs on the amplified Liver data set (1000 samples). We can see that as the hypothesis feature space becomes more complex (i.e., $L$ grows), the training time consumed by both ELM (Gaussian kernel) and GNN increases slowly.
}
\label{fig.Neuron_Time_exp}
\end{figure}

The results of comparison on the computational efficiency, has been shown in Fig. (\ref{fig.Neuron_Time_exp}), we can see the training times consumed by both ELM and GNN do not increase remarkably, as $L$ increases. Even more, the training time consumed by GNN grows a bit slower than ELM with a smoother slope.

Based on the above two experiments and the analysis above, it can be seen that the GNN still has extremely fast training speed when the $S$ and $L$ becomes very large. Thus, it can be concluded that the GNN has great computational scalability and efficiency, with regard to the size of training data sets $S$ and the number of neurons $L$.

\subsection{Sensitivity to Hyper Parameters}
\label{subsec.Sensitivity to Hyper Parameters}

Axiomatically, the selections of the hyper parameters are vital to the performance of the classification algorithms. However, if the model is not very sensitive to changes of hyper parameters, i.e., the best generalization performance could be achieved in a wide range of combinations of hyper parameters, the model would have better utility value, since it require less user intervention in real-world implementation. Thus, in this subsection, the ensuing experiments are conducted to test the sensitivity of ELM and GNN to hyper parameters, Even more, the generalization performances of ELM and GNN when $L$ is very small are also explored. To avoid accidental results, each data point is the average values of 100 trials and the ELM's Gaussian parameters and GNN's GPS parameter follow the optimal settings listed in Subsection \ref{subsec.Benchmark Data Sets and Hyper Parameters Selections_5}.

\subsubsection{Sensitivity to Hyper Parameters} The experiments are conducted on a binary problem (Liver data set) and a multi-class problem (Vehicle data set) to show GNN’s sensitivity to the combinations of two hyper parameters (i.e., $\gamma$ and $L$). Similarly, as a comparison, the same experiments on ELM are conducted to test the sensitivity to the combinations of two similar hyper parameters (i.e., $\delta$ and $L$).

\begin{figure}[]

\centering
  \psfrag{L}[c][c][0.85]{$L$}
  \psfrag{c1}[c][c][0.85]{$1/\gamma$}
  \psfrag{cc}[c][c][0.85]{$\delta$}
\subfigure[ELM on Liver Data Set]{\includegraphics[width=0.85\columnwidth]{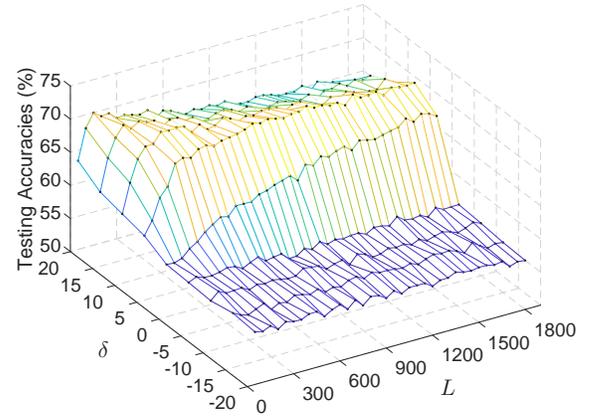}}
\subfigure[GNN on Liver Data Set]{\includegraphics[width=0.85\columnwidth]{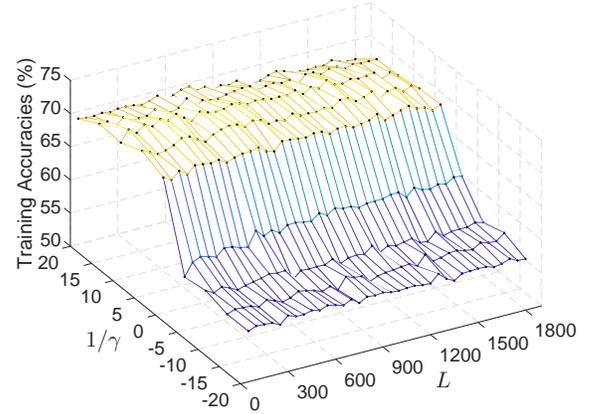}}

\caption{The hyper parametric sensitivity test runs on liver data set (binary). Both ELM and GNN can achieve good generalization performance in a wide
range of combinations of ($\delta$,L) or ($\gamma$,L). However, GNN is less sensitive to $L$ and performs well even when $L$ is small.}
\label{fig.Sensitivity_binary_exp}
\end{figure}

\begin{figure}[]

\centering
  \psfrag{L}[c][c][0.85]{$L$}
  \psfrag{c1}[c][c][0.85]{$1/\gamma$}
  \psfrag{cc}[c][c][0.85]{$\delta$}

  \subfigure[ELM on Vehicle Data Set]{\includegraphics[width=0.85\columnwidth]{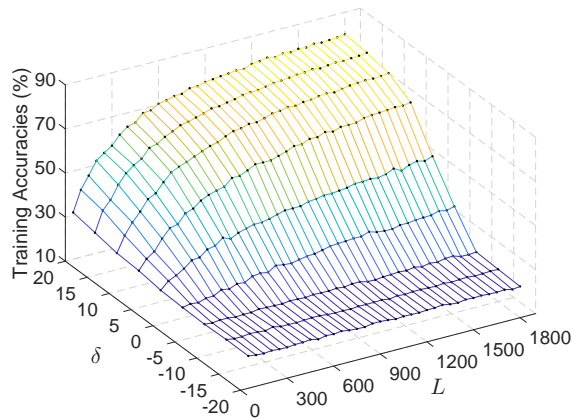}}
\subfigure[GNN on Vehicle Data Set]{\includegraphics[width=0.85\columnwidth]{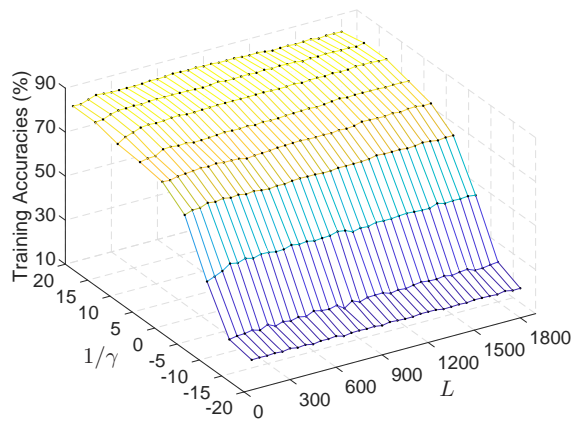}}

\caption{The hyper parametric sensitivity test runs on vehicle data set (mult-class). GNN can achieve good generalization performance in a wider range
of combinations of ($\gamma$,$L$) than ELM. GNN’s performance is only sensitive to the values of  $\gamma$, while ELM could only achieve best performance in a narrow
range of combinations of ($\delta$,$L$).}
\label{fig.Sensitivity_multi_exp}
\end{figure}

According to the results shown in Fig (\ref{fig.Sensitivity_binary_exp}) and Fig. (\ref{fig.Sensitivity_multi_exp}), the best performance of GNN could be achieved in a wider range of combinations of ($\gamma$,$L$). Especially, the performance of GNN is not sensitive to $L$, and can still perform well even when there only exists a very few neurons in the network. Thus, only $\gamma$ need to be specified to achieve best performance. However, the ELM is relatively more sensitive to the combinations of ($\delta$, $L$), and good performance could be achieved only when the $L$ is large enough.


\subsubsection{Performance when $L$ is small} From the Fig (\ref{fig.Sensitivity_binary_exp}) and Fig. (\ref{fig.Sensitivity_multi_exp}), we can also see that when neurons are few, the generalization performance of GNN are still good, while the generalization performance of ELM grows slowly when $L$ grows. Thus, experiments on the same two data sets are conducted to compare the generalization performance of these two algorithms when $L$ is small. The basic experimental settings also follow the ones in Subsection \ref{subsec.Benchmark Data Sets and Hyper Parameters Selections_5}, except ($\delta$,$L$) for ELM and ($\gamma$,$L$) for GNN.

\begin{figure}[]

\centering
  \psfrag{L}[c][c][0.85]{$L$}
  \psfrag{cc=15}[c][c][0.6]{$\delta=15$}
  \psfrag{cc=5}[c][c][0.58]{$\delta=5$}
  \psfrag{cc=0}[c][c][0.58]{$\delta=0$}
  \psfrag{cc=-5}[c][c][0.55]{$\delta=-5$}
  \psfrag{cc=-15}[c][c][0.55]{$\delta=-15$}

  \psfrag{c1=15}[c][c][0.6]{$\gamma=15$}
  \psfrag{c1=5}[c][c][0.58]{$\gamma=5$}
  \psfrag{c1=0}[c][c][0.58]{$\gamma=0$}
  \psfrag{c1=-5}[c][c][0.55]{$\gamma=-5$}
  \psfrag{c1=-15}[c][c][0.55]{$\gamma=-15$}

  \subfigure[ELM on Liver Data Set with Very Few Neurons]{\includegraphics[width=0.85\columnwidth]{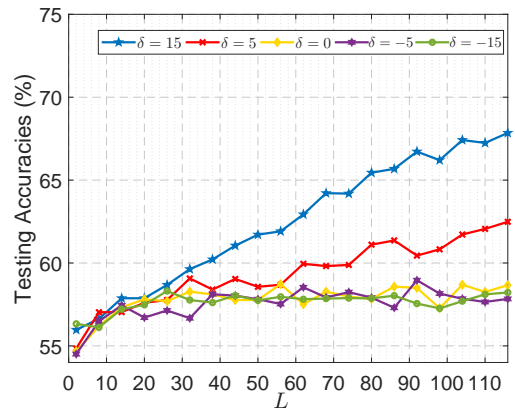}}
\subfigure[GNN on Liver Data Set with Very Few Neurons]{\includegraphics[width=0.85\columnwidth]{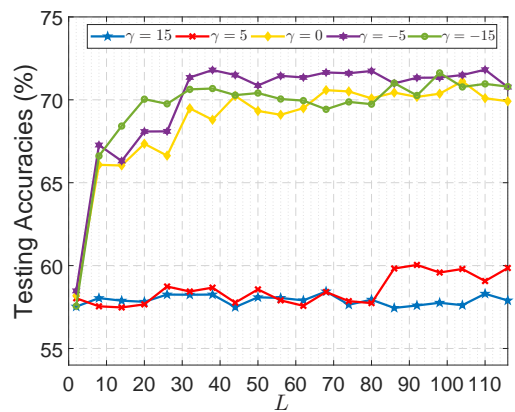}}

\caption{The experiment runs on Liver data set (binary). The ELM starts with a low generalization accuracy when the neurons are very few, and the accuracy grows slowly, while the GNN has relatively high generalization accuracy when $L$ is larger than 10.}
\label{fig.Binary_experiments_with very few neurons}
\end{figure}

\begin{figure}[]

\centering
  \psfrag{L}[c][c][0.85]{$L$}
  \psfrag{cc=15}[c][c][0.6]{$\delta=15$}
  \psfrag{cc=5}[c][c][0.58]{$\delta=5$}
  \psfrag{cc=0}[c][c][0.58]{$\delta=0$}
  \psfrag{cc=-5}[c][c][0.55]{$\delta=-5$}
  \psfrag{cc=-15}[c][c][0.55]{$\delta=-15$}

  \psfrag{c1=15}[c][c][0.6]{$\gamma=15$}
  \psfrag{c1=5}[c][c][0.58]{$\gamma=5$}
  \psfrag{c1=0}[c][c][0.58]{$\gamma=0$}
  \psfrag{c1=-5}[c][c][0.55]{$\gamma=-5$}
  \psfrag{c1=-15}[c][c][0.55]{$\gamma=-15$}

\subfigure[ELM on Vehicle Data Set with very few neurons]{\includegraphics[width=0.85\columnwidth]{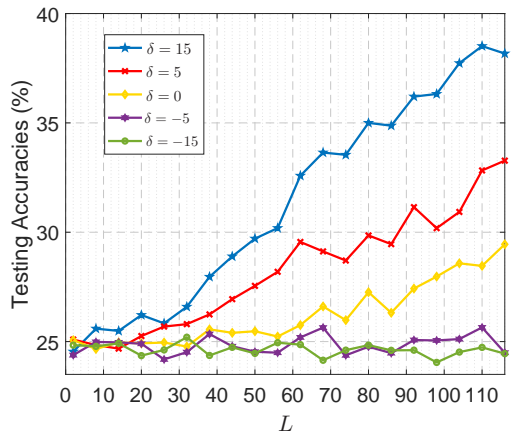}}
\subfigure[GNN on Vehicle Data Set with very few neurons]{\includegraphics[width=0.85\columnwidth]{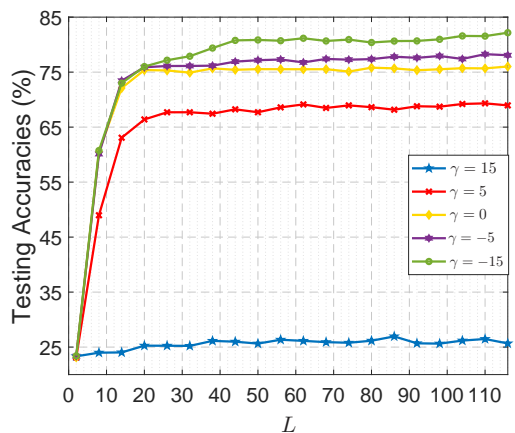}}
\caption{The experiment runs on Vehicle data set (multi-class). The ELM starts with a very low generalization accuracy when the neurons are very few, and the accuracy grows slowly, while the GNN has relatively high generalization accuracy when $L$ is larger than 10. Also, both ELM and GNN are sensitive to the values of $\delta$ or $\gamma$}
\label{fig.Multi_experiments_with very few neurons}
\end{figure}

From the results shown in Fig. (\ref{fig.Binary_experiments_with very few neurons}) and Fig. (\ref{fig.Multi_experiments_with very few neurons}), it is clear that the GNN works still very well with high Accs, when the hypothesis feature space is not complicated (i.e., $L<50$), while for ELM, we can performance deteriorates greatly when lacking enough neurons. Thus, with appropriate values of $\gamma$, the approximation capability of GNN is much better than ELM, which needs more neurons to ensure its approximation capability. As a conclusion, GNN has much better generalization performance than ELM when the network hypothesis feature space is simple (i.e., $L$ is small).

Thus, through the two aforementioned experiments, we can conclude the that the GNN is less sensitive to the changes of hyper parameters than ELM. What is more, it still performs excellently when the dimensionality $L$ of the hypothesis feature space is small. Thus, the model proposed has great utility values and is easy to implement.

%

\begin{table}[]

\scriptsize

\centering

\caption{Performance Comparison of LS-SVM, ELM and GNN: Multi-Class Data Sets with $5\%$ Random Noises}

\label{Tab:Performance Comparison of LS-SVM, ELM and MOGNN: Multi Classification Data Sets with Random Noises1}

\begin{tabular}{cccccccccc}

\toprule

\multirow{2}{*}{Data Sets} & \multicolumn{2}{c}{LS-SVM} & \multicolumn{2}{c}{ELM} & \multicolumn{2}{c}{GNN} \\

\cmidrule(r){2-3} \cmidrule(r){4-5} \cmidrule(r){6-7}

&  Testing      &  Changes

&  Testing       &  Changes

&  Testing       &    Changes   \\

&  Acc (\%)      &   (\%)

&  Acc (\%)      &   (\%)

&  Acc (\%)      &   (\%)    \\

\midrule

Iris  &    94.31     &   -1.91    & 95.21      & -0.83   & \textbf{96.02}   & -0.56  \\

Glass    &63.21    & -5.11  &   63.76      &-4.65 & \textbf{68.55}    & -4.13  \\

Wine&93.81& -2.82  & 96.57   & -1.91  & \textbf{97.77}     & -0.77   \\

Ecoli &80.42   & -5.51    & 82.75  & -4.73  & \textbf{85.41}     & -4.14   \\

Vehicle  &79.21   & -3.98    & 79.51  & -3.65   & \textbf{81.89}    & -2.16  \\
\bottomrule

\end{tabular}

\end{table}

\begin{table}[]

\scriptsize

\centering

\caption{Performance Comparison of LS-SVM, ELM and GNN: Multi-Class Data Sets with $10\%$ Random Noises}

\label{Tab:Performance Comparison of LS-SVM, ELM and MOGNN: Multi Classification Data Sets with Random Noises2}

\begin{tabular}{cccccccccc}

\toprule

\multirow{2}{*}{Data Sets} & \multicolumn{2}{c}{LS-SVM} & \multicolumn{2}{c}{ELM} & \multicolumn{2}{c}{GNN} \\

\cmidrule(r){2-3} \cmidrule(r){4-5} \cmidrule(r){6-7}

&  Testing      &  Changes

&  Testing       &  Changes

&  Testing       &  Changes     \\

&  Acc (\%)      &   (\%)

&  Acc (\%)      &  (\%)

&  Acc (\%)      &  (\%)    \\

\midrule

Iris  &    91.87  &  -4.35  & 92.46    & -3.58    & \textbf{94.59}   & -1.99   \\

Glass    &60.19   & -7.13    &   61.72     & -6.69 & \textbf{66.77}    & -6.23  \\

Wine&92.73  & -4.90    & 95.94   & -2.54    & \textbf{98.21}     & -0.33   \\

Ecoli &78.80   & -7.13    & 80.13    & -7.35        & \textbf{84.51}    & -5.04   \\

Vehicle  &75.27   & -7.92    &77.37    &    -5.79     & \textbf{79.25}     & -4.80 \\
\bottomrule

\end{tabular}

\end{table}

\subsection{Comparison of Classification Robustness}
\label{subsec.Comparison Robustness of Classification}

To investigate the classification robustness of the proposed GNN with R-WDD, we conduct the following experiments on the same real-world benchmark multi-classification data sets as Subsection \ref{subsec.Performance Comparison on Real-World Benchmark Data Sets}. Then, we contaminate the data sets by adding random noises with amplitude of  $5\%$,and $10\%$ to the training sets. And the test set should be noise free since this experiment is to evaluate the classification robustness of the algorithms when noises exist. All experiment settings follow the settings in Subsection \ref{subsec.Benchmark Data Sets and Hyper Parameters Selections_5}.

TABLE \ref{Tab:Performance Comparison of LS-SVM, ELM and MOGNN: Multi Classification Data Sets with Random Noises1} and \ref{Tab:Performance Comparison of LS-SVM, ELM and MOGNN: Multi Classification Data Sets with Random Noises2} show the Accs of three algorithms on different data sets when the noises amplitude is $5\%$ and $10\%$ respectively. Also, we compare the results in TABLE \ref{Tab:Performance Comparison of LS-SVM, ELM and MOGNN: Multiclassification Data Sets} with TABLE \ref{Tab:Performance Comparison of LS-SVM, ELM and MOGNN: Multi Classification Data Sets with Random Noises1}-\ref{Tab:Performance Comparison of LS-SVM, ELM and MOGNN: Multi Classification Data Sets with Random Noises2}, and calculate the changes in Accs before and after the noises being added, to show the influences on the performance of algorithms brought by random noises.

All the performances of algorithms deteriorate as the random noises are added. However, the decreases in Accs of GNN are much slower than that of the LS-SVM and ELM. To be illustrative, we take the Wine data sets as example.
\begin{itemize}
  \item For Wine data set, when the amplitude of noises is $5\%$, the decrease of GNN in Acc is only 0.77\%, while that of LS-SVM and ELM are 2.82\% and 1.91\% respectively; when the amplitude grows to $10\%$  the decrease of LS-SVM and ELM in Acc are 4.90\% and 2.54\%, while that of the GNN is only 0.33\%, 1/15 and 1/8 of the decrease of LS-SVM and ELM. Besides, the Acc of GNN is at least 2\% higher than LS-SVM and ELM when amplitude of noises is 10\%.
\end{itemize}

Based on the above analysis, it could be concluded that the classification robustness of GNN with R-WDD is higher than LS-SVM and ELM.

\section{Conclusion}
\label{sec.conclusion}
In this paper, to tackle the slow operations of BP-type algorithms and computational robustness problems, a novel type GNN model with structural simplicity has been proposed and investigated, based on GPS and polynomials function approximation. Then, in order to overcome the disadvantages of easy over-fitting and improve model's generalization performance, a R-WDD method, based on equality-constrained optimization, has been proposed to minimize networks' empirical risk and structural risk. For better computational scalability and efficiency, two solutions to the R-WDD optimization are given, depending on the scale of the problems and the complexity of the model.

As verifications, the comparison experiments on various real-world benchmark data sets have proved GNN has the comparable (or better) generalization performance, especially in multi classification problems, compared with LS-SVM, ELM. Besides, the excellent computational scalability and efficiency of GNN have been confirmed in the comparison experiments. As for the hyper-parametric experiments, less sensitivity of GNN to the combinations of hyper parameters has been observed. Especially, the generalization ability of GNN hardly relies on the values of $L$ and good performances could be achieved even though the dimensionality of the hypothesis feature space $L$ is very small. Finally, the experiments on contaminated data sets also verified the classification robustness of the proposed method.

\bibliographystyle{ieeetr}
\bibliography{biblio}

\end{document}